\documentclass[10pt,letterpaper]{article}

\usepackage{subcaption}

\usepackage[top=0.85in,left=2.75in,footskip=0.75in]{geometry}

\usepackage{amsmath,amssymb}

\usepackage{ifsym}
\usepackage{changepage}

\usepackage[utf8x]{inputenc}

\usepackage{textcomp,marvosym}

\usepackage{cite}

\usepackage{nameref,hyperref}

\usepackage[right]{lineno}

\usepackage{microtype}
\DisableLigatures[f]{encoding = *, family = * }

\usepackage[table]{xcolor}

\usepackage{array}

\usepackage{comment}
\newcolumntype{+}{!{\vrule width 2pt}}

\newlength\savedwidth

\usepackage{svg} 

\usepackage[aboveskip=1pt,labelfont=bf,labelsep=period,justification=justified,singlelinecheck=off]{caption}

\makeatletter
\renewcommand{\@biblabel}[1]{\quad#1.}
\makeatother

\usepackage{lastpage,fancyhdr,graphicx}
\usepackage{epstopdf}
\fancyheadoffset[L]{2.25in}
\fancyfootoffset[L]{2.25in}
\newcommand*{\ie}{\emph{ie.,~}}

\begin{document}
\vspace*{0.2in}

\begin{flushleft}
{\Large
\textbf\newline{Symmetry perception by deep networks:\\ Inadequacy of feed-forward architectures and\\ improvements with recurrent connections} 
}
\newline
\\
Shobhita Sundaram\textsuperscript{1,$\star$,\Letter},
Darius Sinha\textsuperscript{2,$\star$},
Matthew Groth\textsuperscript{1},\\
Tomotake Sasaki\textsuperscript{3},
Xavier Boix\textsuperscript{1,\Letter}
\\
\bigskip
\textbf{1} Department of Brain and Cognitive Sciences, Massachusetts Institute of Technology, Cambridge, MA
\\
\textbf{2} Buckingham Browne \& Nichols School, Cambridge, MA
\\
\textbf{3} Fujitsu Limited, Kawasaki, Japan
\\

\bigskip

%
%
$\star$ These authors  contributed equally to this work.



\Letter ~Correspondence to shobhita@mit.edu, xboix@mit.edu

\end{flushleft}

\section*{Abstract}

Symmetry is omnipresent in nature and perceived by the visual system of many species, as it facilitates detecting ecologically important classes of objects in our environment.  Symmetry perception requires abstraction of long-range spatial dependencies between image regions, and its underlying neural mechanisms remain elusive. In this paper, we evaluate Deep Neural Network (DNN) architectures on the task of learning symmetry perception from examples. We demonstrate that feed-forward DNNs that excel at modelling human performance on object recognition tasks, are unable to acquire a general notion of symmetry. This is the case even when the DNNs are architected to capture long-range spatial dependencies, such as through `dilated' convolutions and the recently introduced `transformers' design. By contrast, we find that recurrent architectures are capable of learning to perceive symmetry by decomposing the long-range spatial dependencies into a sequence of local operations, that are reusable for novel images. These results suggest that recurrent connections likely play an important role in symmetry perception in artificial systems, and possibly, biological ones too.

\section*{Author summary}
Several animal visual systems, including those of bees, birds and human infants, are able to classify images as symmetric or non-symmetric. This ability is notably different from conventional object recognition classification because it requires addressing an abstract relationship which involves non-local comparisons between image regions. Here, we investigate whether deep neural networks that have proven effective at learning to classify diverse sets of patterns, are able to acquire the symmetric/non-symmetric distinction. Feed-forward deep nets prove quite limited at learning a robust concept of symmetry that can generalize to different image types. However, these limitations can be alleviated, to an extent, by incorporating recurrence in the networks which facilitate identifying symmetric patters via a sequence of local operations that are easier to learn by the network. Taken together, these results indicate that the perception of symmetry in artificial neural networks depends strongly on recurrent connections, and thus provide clues regarding the nature of computations involved in the perception of symmetry in biological networks.


\section*{Introduction}

We inhabit a world wherein several entities that carry great ecological significance for us are bilaterally symmetric \cite{MartindaleHenry, ball2009, manuel2009, Rosen2009}. This includes faces, bodies, animals, and fruits, among many others. The genetic plans of many organisms define symmetric morphologies \cite{davidson2006}. The prevalence of symmetric structures in the natural world is complemented by the exquisite sensitivity humans exhibit in detecting such patterns \cite{Pornstein1985, Wenderoth1994}, a fact that has long been noted by many researchers including Mach~\cite{mach1886} and the Gestalt psychologists \cite{Wertheimer1923, Koffka1935}. Humans can discriminate symmetric from non-symmetric patterns even when they are presented tachistoscopically for a fraction of a second, and efficiently search for symmetric patterns amongst non-symmetric distractors \cite{Wolfe1992, Norcia2002}. Symmetry is an important determinant of the aesthetic rating we assign to a visual pattern, whether that is an abstract structure like the view through a kaleidoscope, or the physiognomy of a person’s face \cite{Scheib1999}. Sensitivity to bilateral symmetry is not unique to humans, but is widespread across the animal kingdom, manifest even in insects and birds \cite{Delius1982, Giurfa1996, Moller1998, Benard2006}. These studies have suggested that symmetry perception may also be an important part of 3D object recognition \cite{mascalzoni2012symmetry}.

The learning of symmetry is particularly interesting in that it represents the acquisition of an abstraction. There is no particular local structure that signals the presence or absence of symmetry; two images with very different pixel compositions can both be members of the ‘symmetric’ class. The criterion that confers class membership is the existence of a relationship between image regions, without imposing any requirements on the contents of the regions themselves. Furthermore, the classification of a pattern as symmetric requires the assessment of long-range relationships \cite{Dakin1997, Dakin1998, Saarinen2000}. Both of these aspects of symmetry assessment are quite different from the classification criteria for many other patterns such as faces or cars. The concepts there are compositional; the presence of a few consistent patterns of pixels that have local extent (say, eyes, mouth, skin color for a face) suffice to correctly classify an image with high confidence \cite{Biederman1987, agarwal2004}. 

Given the remarkable perceptual sensitivity we have to symmetric patterns, several neuroscientists and computer vision researchers have sought to model the mechanisms underlying this ability \cite{Cham1994, Latimer1994, Tyler1996, Wagemans1997, Fukushima2006, Poirier2010, Funk2017, George2021}. However, these models were limited have left open the question of whether the networks have acquired a generalizable concept of symmetry that can apply to image classes different from those used in training and/or had sensitivity to symmetry hard-wired in; they did not investigate the possibility of learning the class of symmetric patterns. This is a significant shortcoming since humans and animals come to exhibit such sensitivity without being given an explicit rule for declaring a pattern symmetric; the ability to acquire symmetric pattern classification through limited experience is a key open avenue for modeling efforts.  

The advent of deep networks presents a valuable opportunity in this regard. Deep convolutional networks have exhibited impressive performance on conventional image classification tasks, particularly object recognition \cite{lecun2015deep}. Given that they are comparable with human performance in these settings, the notion of their sharing basic processing strategies in common with humans is gaining currency \cite{khaligh2014deep,yamins2016using}. Deep networks, in other words, are increasingly considered to be reasonable models of the human visual system insofar as object recognition is concerned.

An interesting test of this far-reaching assertion lies in determining whether DNNs are able to learn a rule for detecting symmetry with the same generalization capabilities that humans, or even simpler animals, exhibit. A positive answer would reinforce the claim of DNNs and humans sharing similar representational strategies, while a negative answer would indicate that there may be fundamental differences between the two, notwithstanding the similarities of their performance on conventional classification tasks. More broadly, this exploration will help to determine whether deep networks can learn symmetry, an abstract spatial concept, through exposure to multiple specific exemplars. 

\begin{figure}[t!]
\centering
\includegraphics[width=0.85\linewidth]{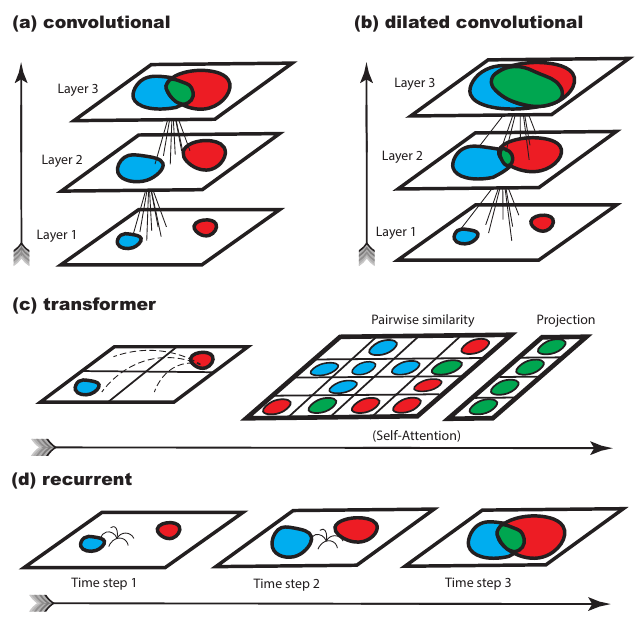}
\caption{\textit{Architectural Mechanisms for Capturing Long-Range Dependencies.} Overview of how the architectures evaluated in this work capture long-range spatial dependencies. We indicate in red and in blue the neurons that are influenced by two different, non-local image regions. Also, we indicate in green the neurons that are influenced by both regions and thus could capture dependencies between them.  (a) The feed-forward convolutional network uses multiple layers to successively expand the receptive field of convolution operations. (b) The dilated convolutional network uses atrous convolutions to expand the receptive field with the same number of parameters. (c) Transformer architectures use self-attention layers to conduct a pairwise comparison of pixel blocks. (d) Recurrent architectures break long-range dependencies into sequences of local operations that are repeated over many time steps (i.e. unrolling steps).}
\label{fig:generalization}
\end{figure}


Given this state of affairs, we conduct two sets of experiments:
\begin{itemize}
    \item EXPERIMENT SET 1: \emph{Symmetry perception by DNNs modelling brain areas for object recognition.} Since brain areas involved in symmetry perception are shared with brain areas for object recognition~\cite{Sasaki2005, Keefe2018}, our first investigation is designed to assess DNNs that have been shown to be effective at modeling the human visual system in terms of object recognition tasks~\cite{khaligh2014deep,yamins2016using}.  

    \item EXPERIMENT SET 2: \emph{Symmetry perception by DNNs with dilated convolutions, recurrence, and transformer networks.} Our second analysis investigates three architecture components that have been designed for problems involving long-range relationships, and are thus especially promising candidates for general symmetry detection. These comprise feed-forward DNNs with dilated convolutions~\cite{dilatedNetwork}, recurrent architectures~\cite{lstmNetwork}, and transformer networks~\cite{Vaswani2017, Radford2021}.
\end{itemize}

We summarize the key differences between the tested architectures across all experiments, and their respective mechanisms for capturing long-range spatial relationships, in Figure \ref{fig:generalization}. To foreshadow the results, we find that only recurrent networks are able to capture long-range relationships and fully generalize out-of-distribution to novel image families. We also report the real-world applicability of these results by training recurrent networks to recognize symmetry in natural images that include background noise and foreground symmetry. In what follows, we describe both these sets of studies and discuss overall inferences from the compiled results.

\section*{\textbf{EXPERIMENT SET 1: Symmetry perception by DNNs modelling brain areas for object recognition}}

We now assess if feed-forward and recurrent DNNs for object recognition, that are deemed as models of human visual processing, are capable of learning a general rule for symmetry detection. This experiment is founded in previous studies showing that symmetry perception activates in brain areas that are shared with object recognition \cite{Sasaki2005, Keefe2018}. In particular, we evaluate the following state-of-the-art models: DenseNet, Xception, InceptionResNetV2, InceptionV3, ResNet101, ResNet50, and RCNN-SAT. To evaluate whether the models have learnt a general rule of symmetry, we test if the DNNs are capable of extrapolating from a limited distribution of synthetic training dataset families.



All images are constructed as matrices of random noise. Random noise is unbiased to particular shapes, allowing us to evaluate recognition of symmetry with less interference from other visual features. 
Image families are differentiated by the presence and size of an uninformative band of pixels at the center of each image. Images with larger central bands place relevant information at the image edges, thus allowing us to evaluate network recognition of long-range relationships. Examples from each dataset family are visualized in Figure \ref{fig:datasets}. To assess if networks have learned a general rule for symmetry detection, and are capable of extrapolating to novel instances, we train on a limited distribution of families (band sizes $0$ and $4$), and test on unseen image families (band sizes $\{2, 4 \text{(dark)}, 6, 14, 16, 18\}$). By testing on images with different band sizes, we ensure that the test images have visual properties that are absent from the training data. 

We examine two training modalities. First, we assess if network representations for object recognition learned from ImageNet already capture symmetry, by transfer-training two fully-connected layers on top of each network with the base layers frozen. We then assess if the networks are capable of learning a generalizable solution by fine-tuning the models end-to-end on our synthetic training families, allowing all layers to update. We find that neither method facilitates learning a general rule for symmetry detection; all trained models struggle to extrapolate to testing families with large band sizes.



\subsection*{Humans easily detect symmetry in our datasets} 
\begin{figure}
    \centering
    \includegraphics[width=0.75\textwidth]{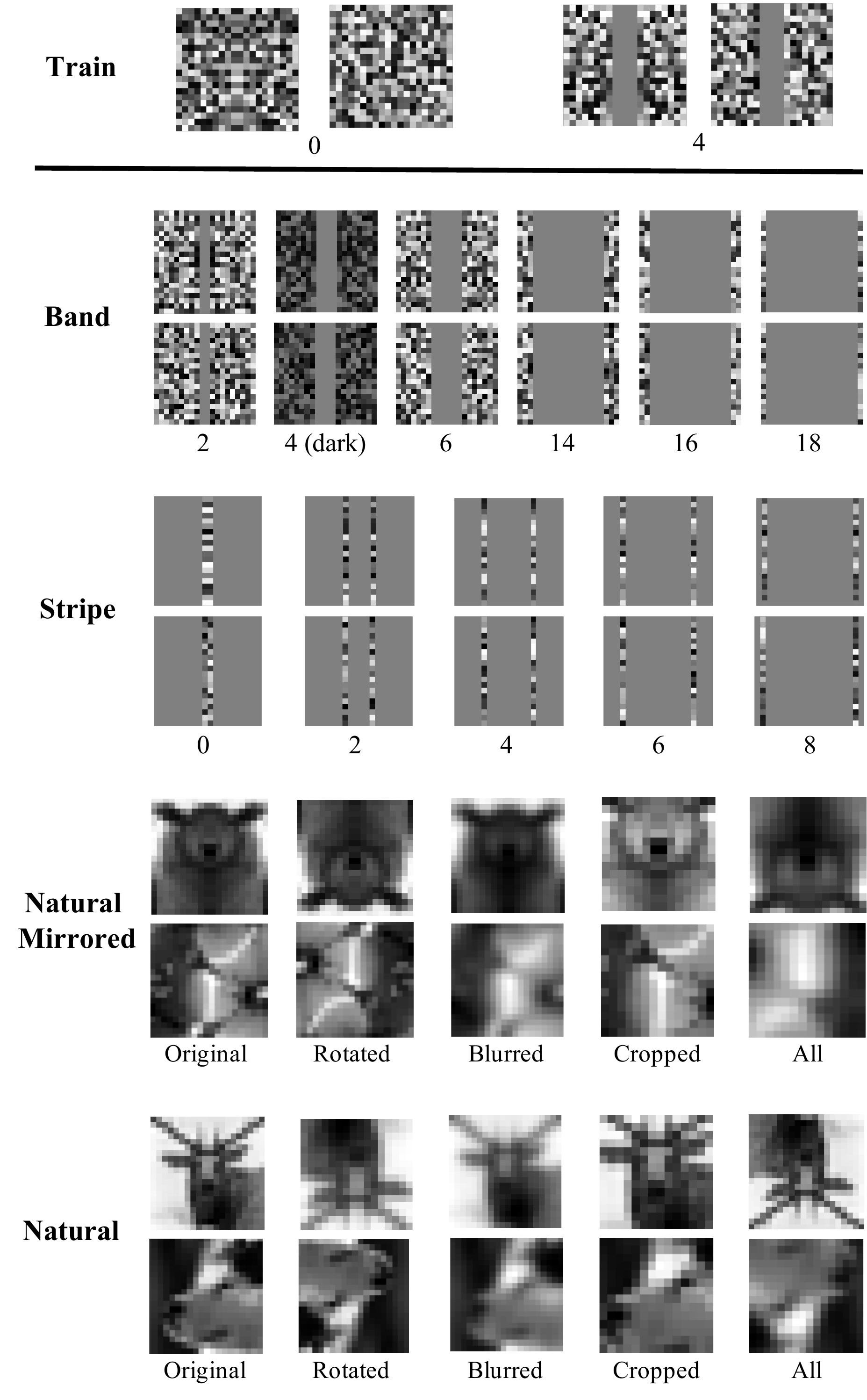}
    \caption{\textit{Datasets for Evaluating Learning of a General Rule of Symmetry Detection.} Examples of synthetic and natural images used to train and test the DNNs. The test set images contain visual properties that do not appear in the training set (i.e. expanded central bands, different distances between flanks), thus enabling generalization testing. Note that in Experiment Set 1 we train only with synthetic images; in Experiment Set 2 we experiment with both synthetic and natural training.}
    \label{fig:datasets}
\end{figure}

Our first study serves as a simple replication of past studies showing rapid learning of symmetry concept. We test the performance of humans in a symmetry classification task. In the process of doing so, we verify that the concept of symmetry is extractable from our designed stimuli. 

For training, ten observers are shown four positive and four negative exemplars with band size 4 (refer to Figure \ref{fig:datasets}). The positive and negative classes are referred to as ‘class 1’ and ‘class 2’, but no explanation is provided as to the class membership criterion, and no mention is made of symmetry. Following this training phase, subjects are shown 50 test images one at a time in five blocks (each block containing five symmetric and five non-symmetric images with band size 4). For each image presented, subjects indicate which class (‘1’ or ‘2’) it belongs to. The image stays on until the participant has responded. No feedback is provided during the test session.
After the initial training with two positive and two negative instances, subjects perform near ceiling when classifying the 50 test images. These results indicate that the symmetry property can be extracted with the stimuli that we have created.


\subsection*{Object recognition DNNs fail to generalize}

\begin{figure}
\centering
\includegraphics[width=\linewidth]{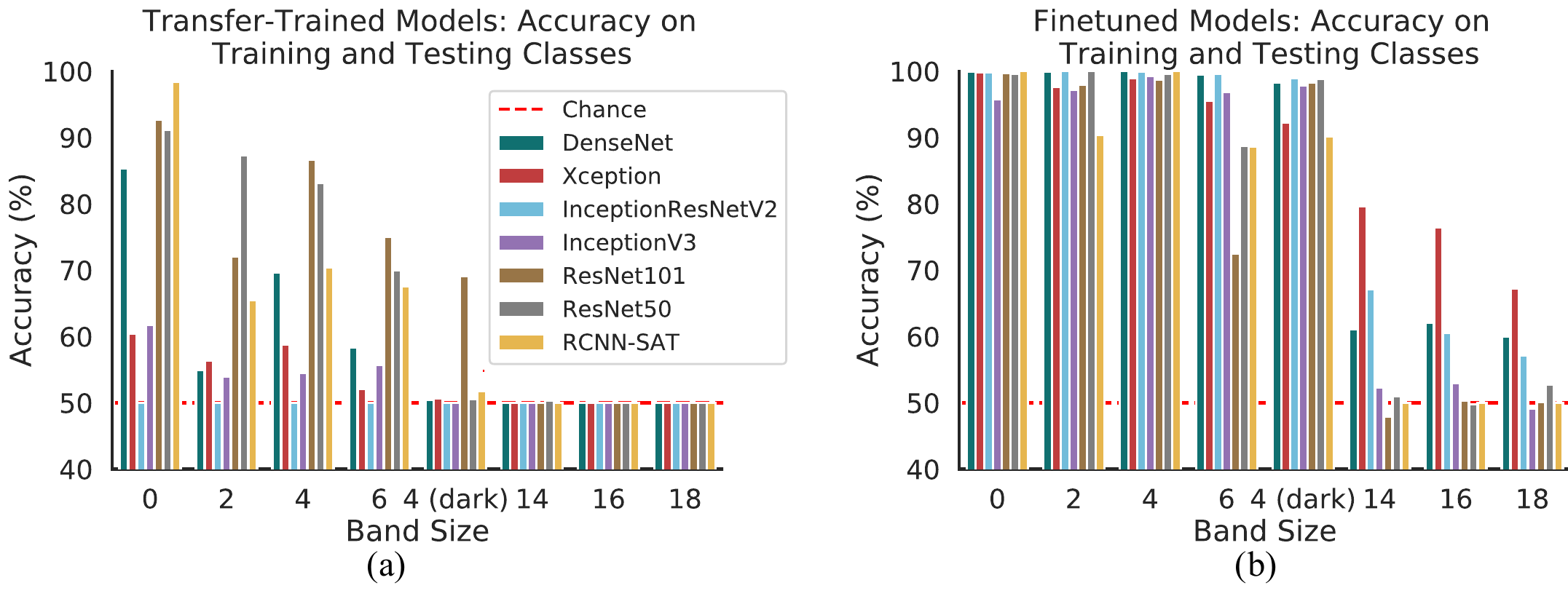}
\caption{\textit{Generalization Performance of Object Recognition DNNs.}  (a) Accuracy for six transfer-trained DNNs (with frozen base models) when classifying new exemplars from training classes or instances of slightly different classes. (b) Accuracy for six pre-trained DNNs trained with end-to-end fine-tuning when classifying the same exemplars. For both (b) and (c) the DNNs do not generalize the categories with large $\{14, 16, 18\}$px band sizes.}
\label{fig:exp1results}
\end{figure}

\paragraph{Transfer-training from ImageNet representations.} We first assess if object recognition representations learned from ImageNet already capture a generalizable notion of symmetry. 
For each network we freeze the base pre-trained layers and train a classification head for the symmetry detection task. To assess generalization we train on symmetric and non-symmetric images with a limited subset of band sizes $\{0, 4\}$ and test on images with band sizes $\{2, 4 \text{ (dark)}, 6, 14, 16, 18\}$ (hyper-parameters are tuned on the validation set of the training families in order to guarantee that the testing families are not used in any way for training, refer to Methods for details). The results are shown in Figure \ref{fig:exp1results}b. We observe that the networks perform poorly across all dataset families, despite showing convergence in training. Performance on novel families is consistently lower than performance on families seen in training. We conclude that the learned representations for object recognition, for both feed-forward and recurrent networks, do not adequately extract features that are relevant to symmetry.

\paragraph{End-to-end fine-tuning on synthetic images.} 

Note that as our DNNs are pre-trained on ImageNet (natural images), the base models may be unable to extract meaningful representations from our synthetic images without further training. Thus, in the following, we allow the networks to update end-to-end, and examine whether they learn a general rule for symmetry detection from the training distribution. 

We train and test each network on the same respective datasets described in the previous section; our results are shown in Figure \ref{fig:exp1results}c. All of the networks demonstrate much stronger performance across the testing families, however are still unable to generalize. Observe that DenseNet, InceptionResNet, and InceptionV3 achieve ceiling accuracy in the training families (band sizes $0$ and $4$) and are capable of generalizing to some testing families (band sizes \{$2$, $4$ (dark), $6$\}). However all networks perform poorly as the band size expands. For band sizes $14$, $16$, and $18$, for which relevant information is localized to the edges of the images, performance drops by $20-50\%$. We conclude that the networks learn some representation of symmetry, but do not fully capture long-range relationships. 

\begin{figure}[t!]
\centering
\includegraphics[width=1\linewidth]{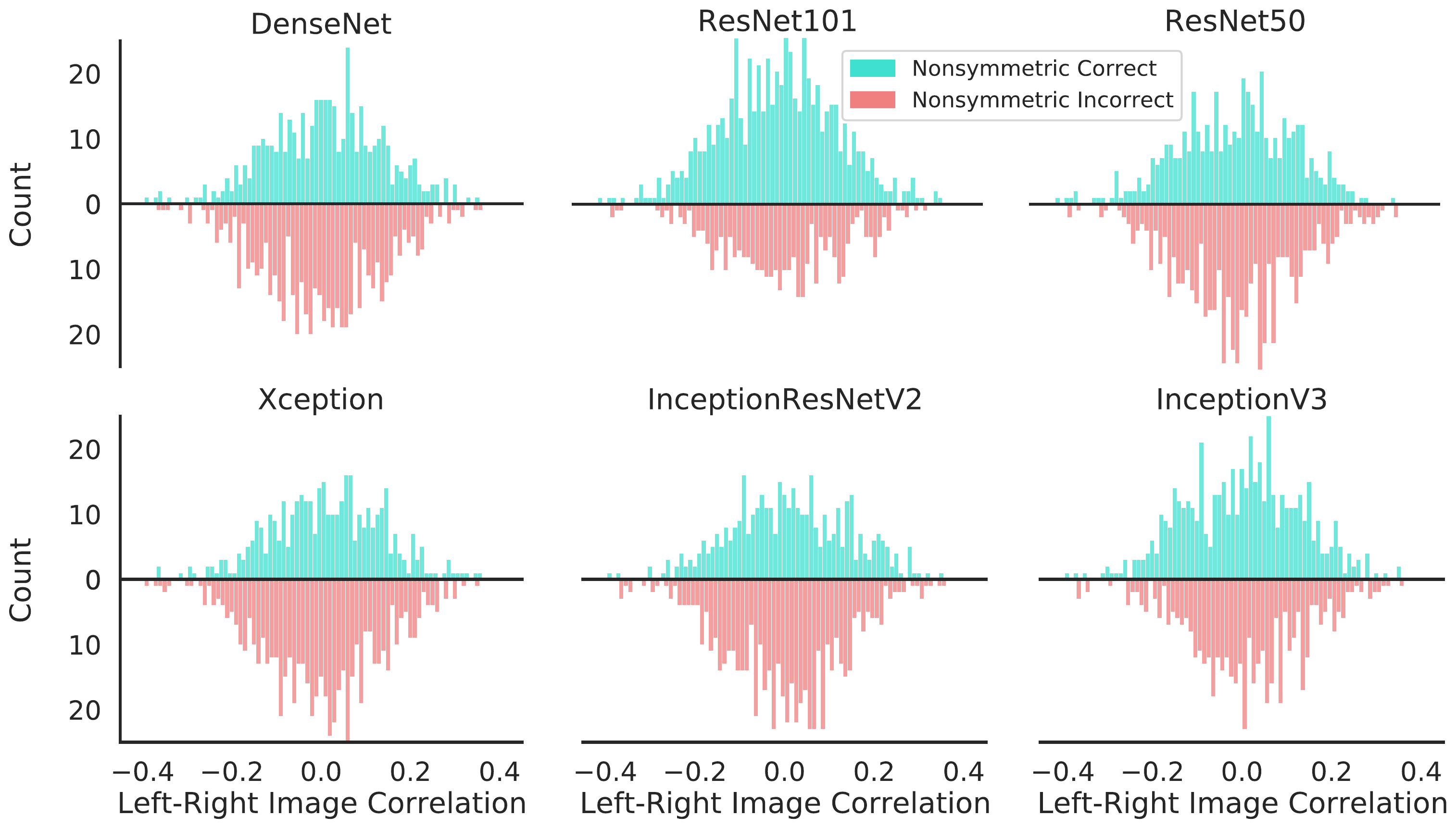}
\vspace{0.1cm}
\caption{\textit{Analysis of Misclassifications Depending on the Degree of Non-symmetry. } We assess the performance of the DNNs on non-symmetric images depending on their degree of non-symmetry. To obtain the degree of non-symmetry, we evaluate the correlation between left and (flipped) right image flanks. We plot the number of images correctly or incorrectly classified depending on such degree of non-symmetry, for images of band size $14$. Results indicates that for all DNNs, the probability score assigned to a non-symmetric image is independent of the degree of non-symmetry.}
\label{fig:histograms}
\end{figure}

\paragraph{Analysis of misclassifications.}
We next analyze the outputs
of the object recognition DNNs to further understand the lack of generalization. In Figure \ref{fig:histograms} we examine the accuracy of the networks depending on the degree of non-symmetry of the images. We assess the degree of non-symmetry as the correlation between left and right image flanks (i.e. how close to symmetric an image is).  We run this analysis on the  fine-tuned networks on non-symmetric images with band size $14$,  as a representative failure mode of the models. If the networks had learned a rule for symmetry detection, we might expect that incorrectly classified non-symmetric images would tend to have a lower degree of non-symmetry. However in Figure \ref{fig:histograms} we see that this is not the case. For all fine-tuned networks, there is no discernible difference in distribution of the degree of non-symmetry (i.e., left-right correlations) for correctly and incorrectly classified non-symmetric images.

\begin{figure}[tp]
\centering
\begin{tabular}{c}
\includegraphics[width=0.9\linewidth]{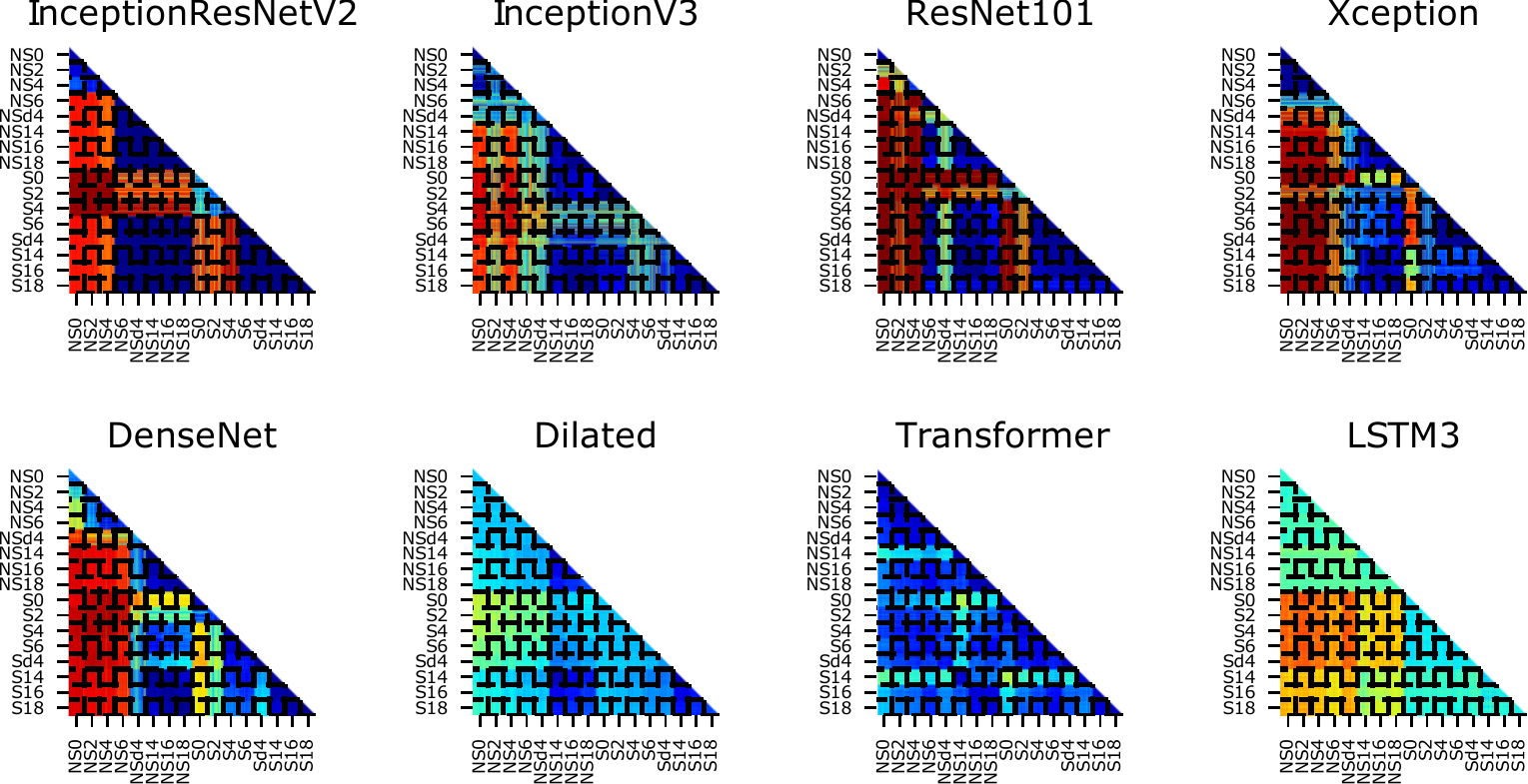}\\
(a) \\[1.5em]
\includegraphics[width=0.65\linewidth]{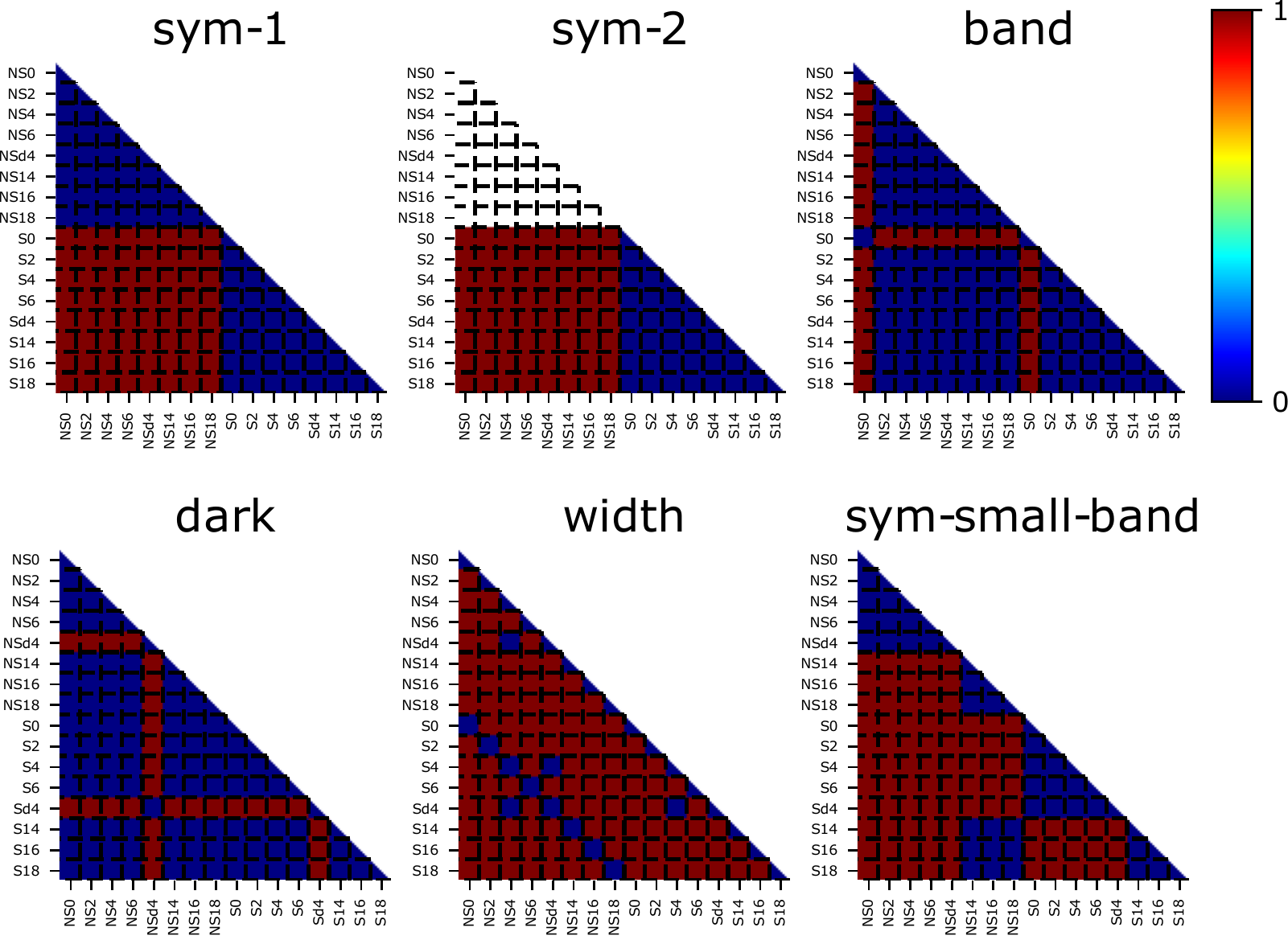}\\
(b) \\[1.5em]
Pearson Correlation between DNN's RDM and Prototypes' RDM\\
\includegraphics[width=0.65\linewidth]{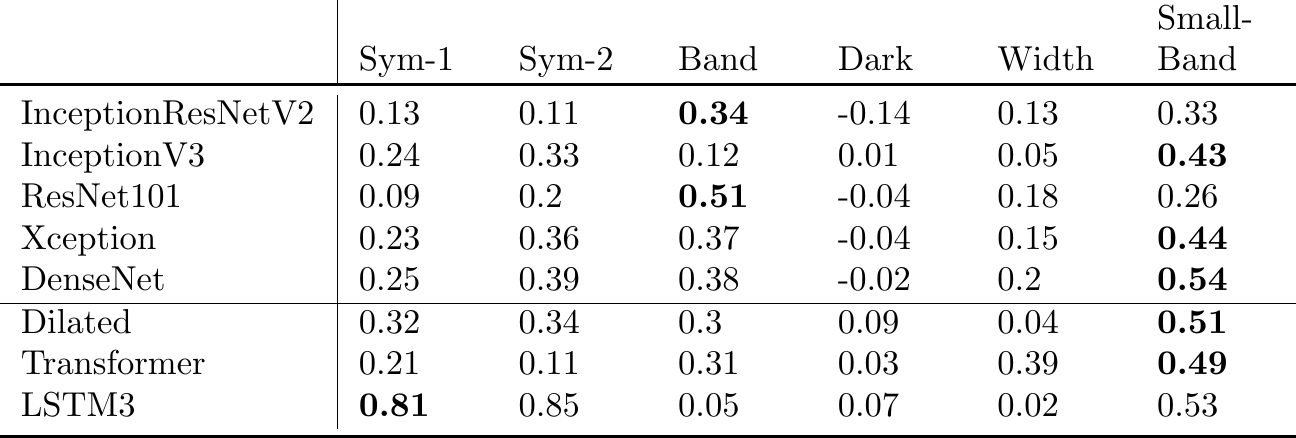} \\
 (c)\\ [0.5em]
\end{tabular}
\caption{\textit{Results of Representational Similarity Analysis (RSA).} (a) RDM of different networks from Experiment Set 1 and 2. (b) Hypothetical patterns showing how the RDMs would look like from perfect classifiers of different features (symmetry, presence of a band, brightness level, or band width). The white area in Symmetry-2 is not included in any correlation calculations. 
(c) Pearson Correlation Coefficients between the RDM of each model and the RDM of each prototype. }
\label{fig:natural_corr}
\end{figure}

\paragraph{Representational Similarity Analysis (RSA).}







To gain insight regarding the image features learned by fine-tuned DNNs to assess image symmetry, we perform an RSA of the ReLU layer before the network's output~\cite{kriegeskorte2008representational}.

Representations  are extracted from $500$ symmetric and non-symmetric images for each of the 8 datasets. For every possible pair of images the cosine distance is calculated, ~\ie $1- \frac{1}{\|\mathbf{x}\|_2 \|\mathbf{y}\|_2} (\mathbf{x}^T \cdot \mathbf{y})$, and then rescaled to $[0, 1]$ to score the dissimilarity between the respective representations. The dissimilarities are displayed in a representational dissimilarity matrix (RDM) in which the images are grouped into the 16 families. Recall that each family contains images that are either symmetric or non-symmetric, with a particular band size and brightness. Thus, the RDM highlights the degree of homogeneity of the representation for each of these features, as depicted in Figure~\ref{fig:natural_corr}a.  

Prototypical models of the RDM are also created to compare our tested DNNs with hypothetical ideal classifiers for each image feature. These prototypical models have assigned dissimilarity scores of either 1 or 0 depending on the feature they classify, as depicted in Figure~\ref{fig:natural_corr}b.  Symmetry-2 is identical to Symmetry-1, except it does not specify the degree of homogeneity when comparing non-symmetric images to other non-symmetric images. This alternative model only needs a strong representation of symmetry, not asymmetry. It may perform just as well in classification tasks. Symmetry-Small-Band is a model which fails with large bands. Thus, the RDM is a function of whether images are symmetric only when comparing images that both have small bands. 

We validate these qualitative observations by gathering Pearson correlation coefficients between the RDMs of each neural network model and each prototypical model. The correlation results can be seen in Figure~\ref{fig:natural_corr}c. These analyses also include the \textit{LSTM}, \textit{Dilated}, and \textit{Transformer} models, which are introduced and analyzed in Experiment Set 2. The Symmetry-Small-Band prototype has the highest correlation with the majority of networks investigated in Experiment Set 1. Band-Presence has the highest correlation with InceptionResNetV2 and ResNet101. Thus, the  DNNs models tested in this experiment rely on the presence of the band rather than on a general rule of symmetry. 

\subsection*{Discussion}
Our results indicate that state-of-the-art  DNNs that have shown impressive performance on object recognition tasks, find it challenging to learn a rule for bilateral symmetry detection that can then be applied to images with different visual contents. In our experiments, the networks struggle to extrapolate to images with larger band sizes when trained on images with a subset of possible band sizes. This limitation holds for networks trained both with and without end-to-end fine-tuning.   Note that the accuracy drops as the band size increases, indicating that the DNNs do not learn to effectively evaluate long-range relationships. Therefore, it is plausible that our networks have taken shortcuts in the learning process to most efficiently and accurately distinguish between symmetric and non-symmetric images in our training dataset, but failed to extrapolate a general concept of symmetry.  

These results are consistent with previous studies demonstrating that object recognition networks struggle to generalize beyond the training distribution, such as object recognition with out-of-distribution object orientations \cite{Alcorn2019,madan2022}. Our study adds to the recent body of works that have suggested that DNNs fail to learn general solutions especially when long-range dependencies and abstract concepts are involved~\cite{insideness}. 


The limited generalization capabilities of the networks may be attributed to the fact that the networks were initially trained for object recognition and as a result, the representations learned may be unable to encode the long-range dependencies required for symmetry perception. Another possible reason is that the network architectures are insufficient to capture long-range dependencies. Note that we evaluate one network with recurrent connections, i.e., RCNN-SAT, as recurrent architectures are better suited to learning representations that capture long-range dependencies than purely feed-forward architectures. RCNN-SAT performs on par with the feed-forward DNNs, suggesting that pre-training for object recognition may be the main cause of the networks’ limited generalization capabilities. Yet, the RCNN-SAT recurrent architecture may also not be suitable for symmetry perception.

Our results further call into question the adequacy of these systems as models of human visual perception. It is likely that generalization would improve if images with larger band sizes were included in the training data set. However, even if these object recognition networks were to be able to learn symmetry with much more diverse training sets than the ones we used, that would still not entitle them to be considered good models of biological vision, since animals are able to acquire the concept with very few training instances~\cite{Delius1982, Giurfa1996, Benard2006}. 


Given the manifest limitation of object recognition networks to acquire the symmetry concept, we shift our focus from networks that are deemed as models of human object recognition to architectures that could allow the networks to gain sensitivity to long-range spatial relationships. The second set of experiments considers three such architectures.  

\section*{EXPERIMENT SET 2: Symmetry perception by DNNs with dilated convolutions, recurrence, and transformer networks}


In this second set of experiments, we investigate three architecture that may enable capturing long-range dependencies. These are: the Dilated Convolutional Neural Network (\textit{Dilated}) \cite{dilatedNetwork}, a three stacked Convolutional LSTM (\textit{LSTM3}) \cite{lstmNetwork}, and a Transformer \cite{Radford2021}. To our knowledge, none has previously been applied to learning-based symmetry detection. In Figure \ref{fig:generalization} and Methods, we describe how these networks are able to accomplish long-range comparisons.

 We analyze the \textit{Dilated} and \textit{LSTM3} architectures directly trained for symmetry perception, without pre-training to object recognition. This facilitates studying the learning of symmetry in isolation, independent of visual cues related to object recognition that may inhibit symmetry perception. Thus we can effectively evaluate the impact of introducing various architectural components.
The \textit{Transformer} network could not be trained from scratch in symmetry perception, as it requires hundreds of millions of training examples and an inaccessible amount of computational resources. We used the standard transfer-training procedure for Transformers, pre-trained using $400$ million text-image pairs  (see Methods). This procedure has been shown to lead to powerful visual representations capable of successfully performing a plethora of tasks~\cite{Radford2021}. 

We first assess our networks on the same synthetic datasets as the previously-described object recognition models to compare their respective performances.
We then further assess our networks' generalization capabilities by testing them on further novel dataset families (denoted as Stripe datasets) that differ significantly from the training data. In order to fully investigate the generalization capabilities of our networks, we also introduce the Natural Mirrored and Natural datasets, which assess if the networks are capable of learning and detecting symmetry in the natural world ("in the wild"). 

\subsection*{Synthetically-trained LSTM achieves high generalization accuracy}

\paragraph{Cross-dataset testing with synthetic data.} To demonstrate a systematic improvement in generalization, we first evaluate the performance of the \textit{Dilated}, \textit{LSTM3}, and \textit{Transformer} networks on the Band datasets used in Experiment Set 1. Recall that all networks are trained on symmetric and non-symmetric images with band-sizes $0$ and $4$, and tested on images with band-sizes $\{2, 6, 4 \text{(dark)}, 14, 16, 18\}$ to assess generalization accuracy (recall that the hyper-parameters are tuned on the validation set of the training families in order to guarantee that the testing families are not used in any way for training). 

\begin{figure}[t!]
\centering
\includegraphics[width=1\linewidth]{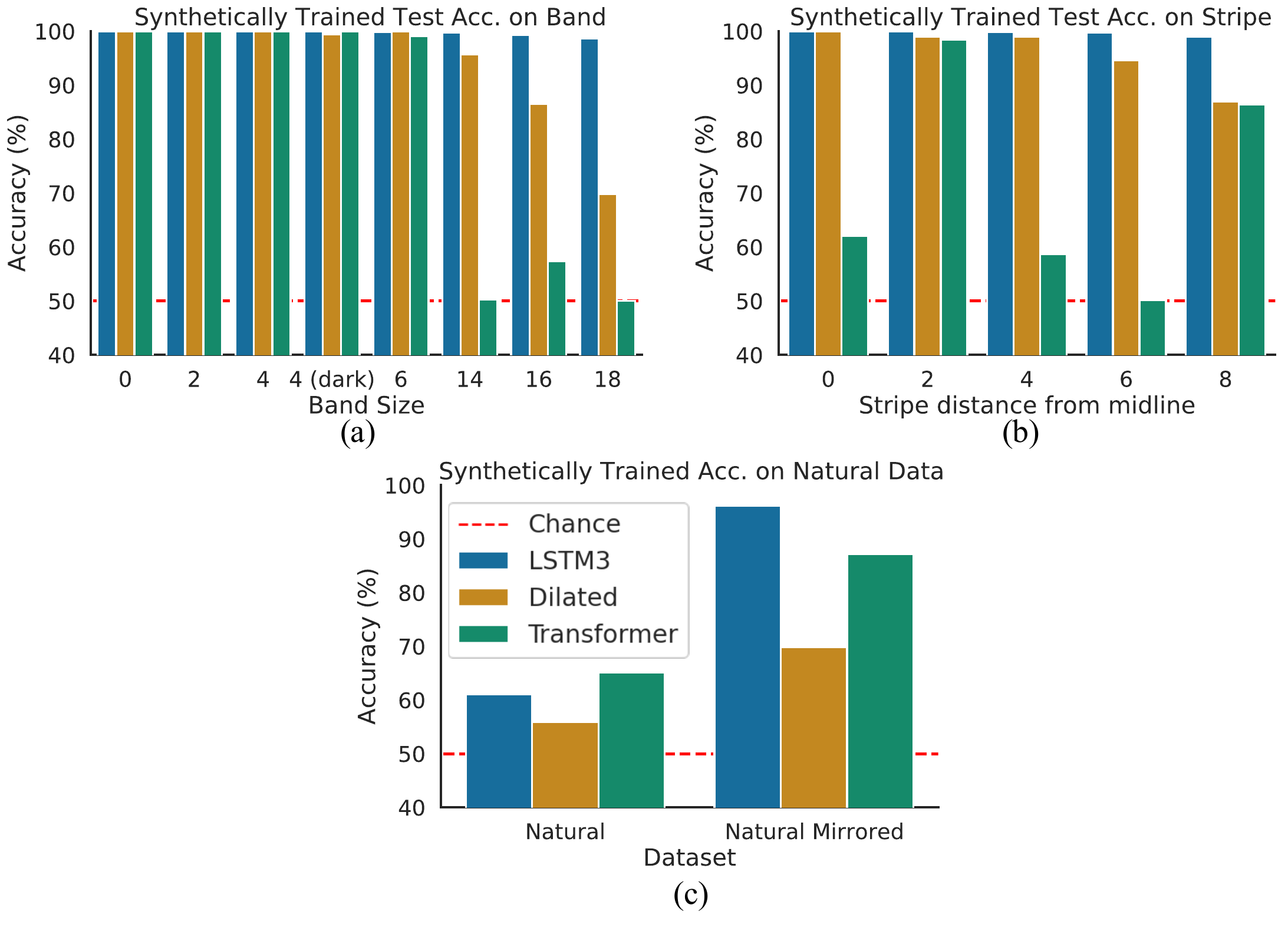}
\caption{\textit{Generalization Performance of Synthetically Trained LSTM3, Dilated, Transformer.} (a) Cross-dataset evaluation accuracies on the Band datasets. (b) Accuracies on the Stripe datasets. (c) Accuracies on natural image datasets. (d) LSTM3 accuracy on Band datasets for different training data sizes. Across all datasets, only the LSTM3 fully generalizes.}
\label{fig:synthetic_acc}
\end{figure}

The accuracies across these categories are shown in Figure \ref{fig:synthetic_acc}a. Recall that the object recognition models achieve relatively poor testing accuracies in images with large central bands despite achieving $100\%$ accuracies in training categories, thus demonstrating overfitting. In striking contrast, our \textit{LSTM3} achieves near-perfect accuracy across both training and testing categories. The \textit{Dilated} and \textit{Transformer} networks demonstrate similar trends as the object recognition networks. The \textit{LSTM3} results demonstrate a massive improvement in generalization accuracy (on the same synthetic datasets) compared to Experiment Set 1.

We next evaluate all three networks on test datasets that differ even more greatly from the training distribution. We test on the Stripe family of datasets, for which each image has just a single column of varying pixels on either flank. The Stripe datasets allow us to further evaluate the relationship between network accuracy and the distance between image flanks, thus examining how well the learned solutions evaluate non-local relationships.

The accuracies of the networks for Stripe images are shown in Figure \ref{fig:synthetic_acc}b. Note that across all of these categories, similarly to the Band datasets, the \textit{Dilated} and \textit{Transformer} networks perform worse on categories where the informative image regions are restricted to the edges of the image. This trend is particularly evident for \textit{Dilated}. We conclude that the \textit{Dilated} network’s accuracy is dependent on the distance between image flanks, despite its expanded receptive field. While the \textit{Transformer} does not exhibit such a clearly interpretable pattern, it is also clearly does not learn a general rule. In contrast, the \textit{LSTM3} achieves near-perfect accuracy across all categories. The performance of the \textit{LSTM3} is essentially invariant to the distance between image flanks, or the location of informative regions of the image, providing further evidence that based on the limited distribution of training data it learns a solution that effectively captures long-range relationships. 


\paragraph{Testing synthetically-trained networks on natural data: pixel-level symmetry and perceived symmetry.}
We next test the synthetically trained \textit{LSTM3}, \textit{Dilated}, and \textit{Transformer} on datasets of natural images, denoted as the Natural Mirrored and Natural test datasets ($1200$ natural images with mirrored left-right flanks and $1200$ fully natural images respectively). The difference between these two datasets is that the Natural Mirrored contains symmetric patterns at the pixel level, as there is a one to one correspondence between pixels at corresponding position of the respective flanks. All datasets previously evaluated in the paper also contain such pixel-level symmetry. In contrast, the Natural dataset evaluates perceived symmetry, as the objects in the image may be symmetric but due to other visual factors (noise, illumination, background, etc.) there may not be a one to one correspondence between pixels at corresponding position of the respective image flanks.

The results are shown in Figure \ref{fig:synthetic_acc}c. For Natural Mirrored images –which are symmetric at the pixel level, as all previous datasets-- \textit{LSTM3} achieves near perfect ($97$\%) accuracy while \textit{Dilated}  and \textit{Transformer} performs worse, with $70$\%  and $89\%$ accuracy, respectively. These results further strengthens  the evidence  that the synthetically-trained \textit{LSTM3} captures a solution to detect pixel-level symmetry, as it even generalizes to distributions of pixels found in the natural world. 

All networks perform poorly on the Natural dataset; this result is expected given that the networks were only trained on pixel-level symmetry, while natural images incorporate additional factors that prevents the image flanks from being exactly equal at the pixel-level but that are perceived equal. Later in the paper, we further analyze this phenomenon by training the networks for symmetry detection with natural factors. Next, we analyze the strategies learned by the LSTM network to generalize from a restricted training distribution in pixel-level symmetry detection.

\begin{figure}[t!]
\centering
\includegraphics[width=0.8\linewidth]{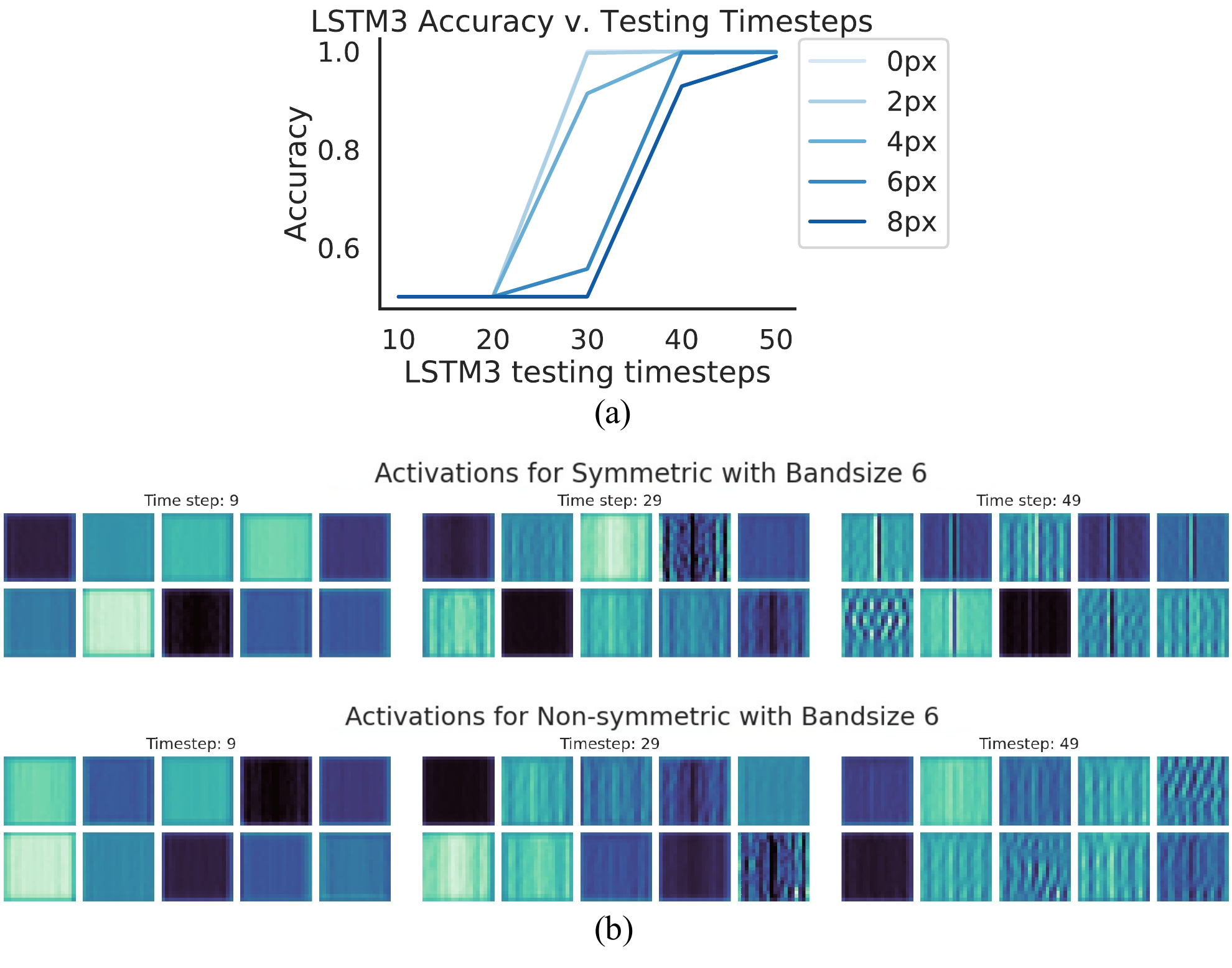}
\caption{\textit{Analysis of LSTM3 Representations.} (a) Plot of the accuracies achieved by the \textit{LSTM3} with restricted numbers of time steps for different categories of Stripe datasets (indicated with the number of pixels between stripes). (b) Visualization of $10$ representative activation maps from the last \textit{LSTM3} cell at various timesteps, for symmetric and non-symmetric images. Each pixel represents the degree of activation of a neuron in the activation map (darker colors mean values closer to 0 while lighter colors mean larger values). Note that as the timesteps progress, the center axis becomes highlighted for symmetric samples.}
\label{fig:mechanism}
\end{figure}

\subsection*{LSTM3 learns a visual routine for solving symmetry}
We now analyze why the synthetically-trained \textit{LSTM3} is the only network that succeeds in capturing long-range relationships. A theoretical explanation is that \textit{Dilated} solely incorporates an expanded receptive field through atrous convolutions, and such large receptive fields may lead to more complex models that overfit. Meanwhile, the pre-trained representations for visual inputs in the \textit{Transformer} may not capture symmetry, thus preventing transfer-training from learning a general solution. In contrast, \textit{LSTM3} not only expands the receptive field size, but is also capable of breaking long-range dependencies into a sequence of local operations. Recall that the architecture of the \textit{LSTM3} involves applying a feed-forward architecture repeatedly over many time steps steps; the unrolled version of the \textit{LSTM3} is a deep network with shared weights across layers. Thus, with recurrent connections we achieve large-receptive fields while controlling network complexity.

\paragraph{Relationship between accuracy and time step.} We examine how the distance between image flanks affects the number of time steps required to correctly classify the image as symmetric or non-symmetric. The presence of such a relationship provides crucial insight regarding the mechanism of the \textit{LSTM3}’s learned solution, indicating that it has learned a visual routine composed of local operations that involves expanding outwards from the image center. 

To elucidate this relationship, we examine the \textit{LSTM3} testing accuracy across the Stripe categories for different numbers of time steps. Recall that each Stripe image has a single column of varying pixels in either flank that is $x \in \{2, 4, 6, 8, 10\}$ pixels from the midline. In Figure \ref{fig:mechanism}a we show the testing accuracy vs. number of time steps for each Stripe category. We observe that when the image flanks are close together (image stripe 0-2 pixels from the midline) the \textit{LSTM3} requires less than $30$ time steps to achieve a high accuracy $>99\%$. When they are the further apart (image stripe $8$ pixels from the midline) 50 timesteps are needed. This result suggests that the \textit{LSTM3} may handle long-range dependencies by propagating information outwards from the center of an image over time.



\paragraph{Representations of symmetric images}
We gain further insight into the mechanism of the \textit{LSTM3} learned solution by visualizing the neural activations. In particular, for symmetric and non-symmetric images with a band size of $6$, we extract the activations of the final \textit{LSTM3} cell. We then perform KMeans clustering on the aggregated activation maps from the same class (with $10$ clusters) to elucidate any common patterns. The KMeans clustering algorithm groups the activation maps together into clusters based on each map's Euclidean distance to the nearest cluster centroid. We choose $10$ because with this number we observe some redundancy between cluster centroids, indicating that we are seeing a representative range of activity. 

The $10$ representative activity maps for symmetric and non-symmetric samples respectively are depicted in Figure \ref{fig:mechanism}b. We depict the clusters for activations extracted at 3 different \textit{LSTM3} time steps. Note that the activations taken at the last time step are considered the "output" activations used to devide whether the image is symmetric or not. We observe that at time step 9, the symmetric and non-symmetric activation maps are indistinguishable. At time step 29 the center regions are highlighted more. At the final time step the midline emerges as the primary highlighted region for symmetric centroids, while for non-symmetric centroids the representation patterns propagate uniformly across the whole map. For symmetric samples, by the final time step the centroids are visually homogenous whereas for non-symmetric images the activations do not demonstrate any such pattern. These observations suggest that the identification of the axis of symmetry is important for the \textit{LSTM3}, and support the hypothesis that the network propagates from the center.

These visualizations affirm key differences in how the \textit{LSTM3} represents symmetric and non-symmetric images, and provides further evidence that recurrent connections allow the network to handle long-range dependencies by propagating information over many time steps. 

\paragraph*{Representation Similarity Analysis (RSA).}
In Experiment Set 1, we introduce the RSA analysis (Figure~\ref{fig:natural_corr}). We find that most networks except \textit{LSTM3}, have the highest correlation with prototypical classifiers of band presence or symmetry only at small band sizes. In contrast, \textit{LSTM3} has a representation that correlates most with a prototypical symmetry classifier that fails with large band sizes (Symmetry-2).

\begin{figure}[t!]
\centering
\includegraphics[width=1.0\linewidth]{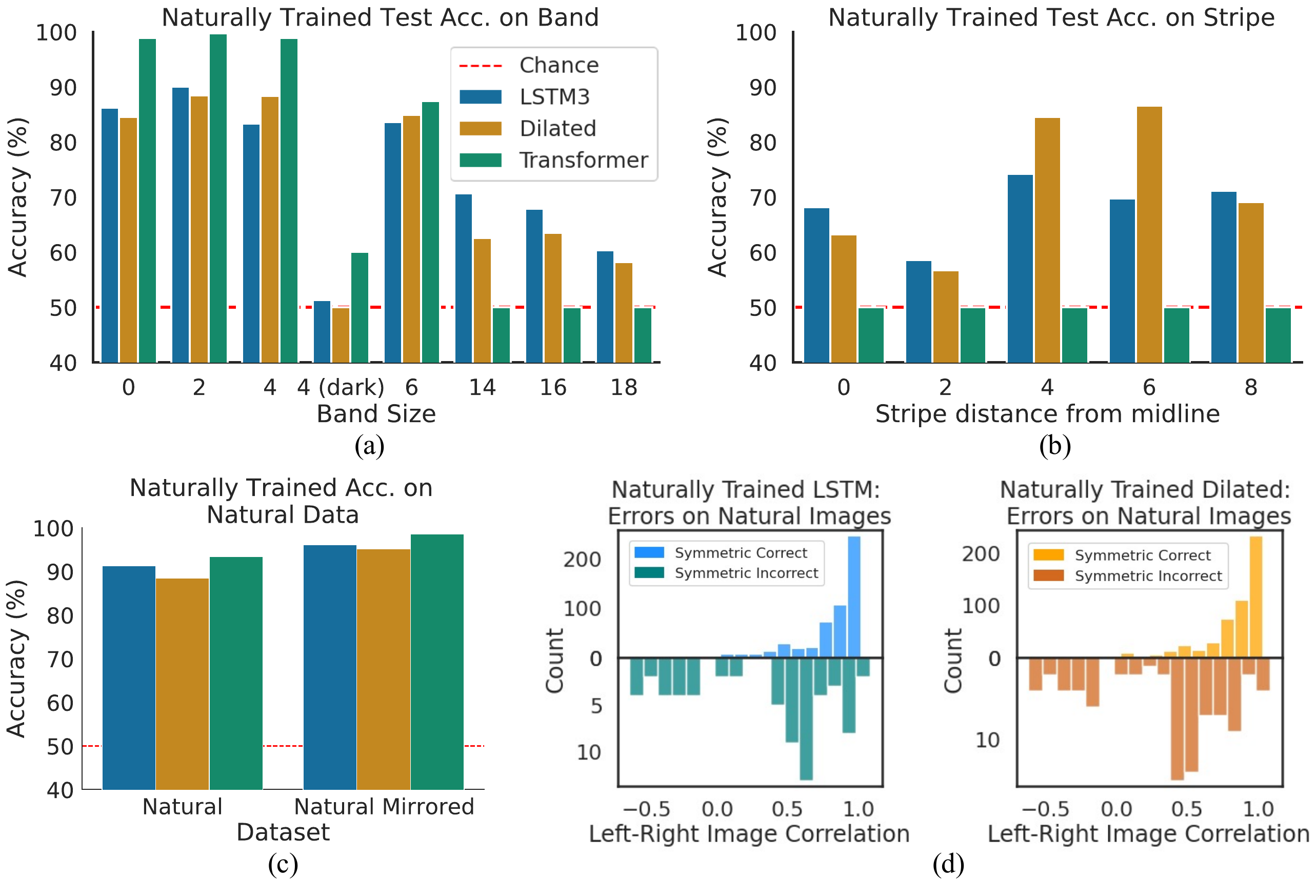}
\caption{\textit{Generalization Performance of Naturally Trained LSTM3, Dilated, Transformer.} (a) Accuracies on the Band datasets. (b) Accuracies on the Stripe datasets. (c) Accuracies on natural image datasets. (d) An analysis of symmetric images correctly/incorrectly classified by \textit{LSTM3} and \textit{Dilated}. The histograms show the number of images correctly or incorrectly classifoed depending on the degree of non-symmetry, which is evaluated as the correlation between the left and right (flipped) halves of the images, where a higher correlation indicates that an image is more symmetric.}
\label{fig:natural_acc}
\end{figure}

\subsection*{Introducing natural factors inhibits symmetry perception}
Finally, we investigate if the \textit{LSTM3}, \textit{Dilated}, and \textit{Transformer} networks are capable of learning to capturing symmetry in natural images. This task is more complex due to the additional factors (such as noise, background, different illuminations in different parts of the image, etc.) present in natural images that may interfere with the detection of symmetry. Thus, this task requires to acquire a more general notion of symmetry than a pixel-level one, as other visual factors may be present in the image. 

We train the \textit{LSTM3}, \textit{Dilated}, and \textit{Transformer} networks on $10800$ augmented natural images, using the same learning and hyper-parameter search procedure as synthetic-training experiments (refer to Methods). We then perform cross-dataset evaluation, testing the naturally-trained networks with the highest validation accuracies on all of the test datasets (Band, Stripe, Natural Mirrored, Natural). 

In Figure~\ref{fig:natural_acc}c we show the cross-dataset accuracy for the natural-trained networks on natural images, and observe that the \textit{LSTM3}, \textit{Dilated}, and \textit{Transformer} networks achieve high accuracies ($91\%$, $88\%$, and $92\%$ respectively for natural data and $>95\%$ for natural mirrored data). In contrast, as seen in Figure \ref{fig:natural_acc}a-b, all three networks fail to generalize back to the synthetic test sets. Unlike the results for synthetically-trained networks, the \textit{LSTM3} does not demonstrate a significant performance improvement over either \textit{Dilated} or \textit{Transformer}. Both the \textit{LSTM3} and \textit{Dilated} networks appear to gain some notion of symmetry, performing well above chance accuracy for almost every test dataset, however the full generalization seen with the synthetically-trained \textit{LSTM3} is not evident.


We gain additional insight by examining the accuracy of the network depending on the correlation between left and  right (flipped) image flanks (i.e. the degree of non-symmetry of an image) for symmetric natural images that are correctly and incorrectly classified; the results are shown in Figure \ref{fig:natural_acc}d. The correctly classified images tend to have a higher left-right correlation (i.e. are more symmetric), while incorrectly classified images have a much wider spread of correlations. These results imply that the networks have picked up some notion of symmetry, and that factors that decrease the visual presence of pixel-level symmetry (evident in decreased left-right correlations) can inhibit the networks' performance.

These results are likely due to the external factors in natural images (noise, illumination, etc) that interfere with the presentation of symmetry, thus making it more difficult to learn symmetry perception from these     ``in the wild" images. While all networks perform well for test images that are similar to the training set, they struggle with the synthetic data that requires recognition of purely pixel-level symmetry ---a visual feature that is more difficult to glean from the natural data.

\subsection*{Discussion}

Symmetry perception presents two primary challenges. First, it is an abstract feature, indicated by the relationships between pixels, such that a symmetric image with unfamiliar local content should still be recognizable as symmetric. Second, these relationships can be long-range. These challenges are present for both computational and human visual systems; thus symmetry is an excellent assay for studying computational models of human perception. 


Our experiments show that only the \textit{LSTM3} model is capable of generalizing to detecting mirror symmetry in novel dataset families, particularly those for which informative regions are located at the image edges. Analysis of the network activations indicate that the \textit{LSTM3} may have learned a routine for symmetry detection that decomposes the long-range dependencies in a sequence of local steps. The \textit{Dilated} and \textit{Transformer} networks, in contrast, struggle with image classes that stress long-range dependencies, and images with different pixel distributions (i.e. natural mirrored images).

These results are consistent with existing works that study the importance of recurrence in modelling visual problems that involve long-range dependencies~\cite{linsley2020stable,receptiveFields,insideness}.
Villalobos et al. have also indicated that recurrent architectures are of critical importance to enable generalization beyond the training distribution in problems with long-range dependencies, in particular for the problem of determining the inside and outside of a closed curve~\cite{insideness}. Our results provide further evidence for the importance of recurrence in enabling DNNs for vision to learn generalizable representations of fundamental visual features that involve long-range relationships. Our results carry implications for downstream tasks, as symmetry in particular is a fundamental visual attribute that has been implicated in facilitating tasks such as pose estimation and depth estimation \cite{Zhou2021, Zhang2020}.

Furthermore, a body of works in the  neuroscience literature argues that recurrence is a crucial component for object recognition~\cite{kietzmann2019recurrence,kar2019evidence,clarke2015predicting,freiwald2010functional,brincat2006dynamic}.  Our results demonstrate that recurrent networks enable generalized symmetry perception in a simulated setting. These indicate a possible role for recurrence in human neural processing of long-range dependencies, and in particular symmetry. Such a hypothesis could be experimentally investigated with human EEG readings. If the human brain detects symmetry through a purely feed-forward process, an EEG might take a shorter time to show settled brain activity than if recurrent computations were involved. Our suggested investigation could confirm a  functional role of recurrent connections that has not been previously considered, i.e. in handling long-range dependencies given a restricted distribution of training exemplars.

The superior generalization performance observed for the synthetically trained \textit{LSTM3}, however, is not apparent for the \textit{LSTM3} trained on natural images. The natural-trained \textit{LSTM3} successfully generalizes to natural images from a similar distribution, however fails to generalize back to synthetic data, indicating that it has not learned to generally discern symmetry. We also observe that incorrectly classified symmetric natural images tended to have a lower degree of symmetry. These results indicate that while the networks represent some notion of symmetry, natural factors make learning symmetry more difficult. While learning to perceive symmetry in isolation appears to require recurrence -as evidenced by our synthetic-training experiments- learning symmetry detection from “in the wild” images may require other modes of visual processing to account for background asymmetries, noise, different illuminations, and other such factors. Evidently, more investigation of generalizable detection of symmetry in natural images is needed.  Future work may involve an architecture search for a model that combines the advantages of recurrence with the advantages of an object recognition model. A promising approach could build upon RCNN-SAT~\cite{rcnnSAT}, as RCNN-SAT models to a substantial extent the temporal dynamics of human object recognition with recurrent connections. As suggested by our results, changes in the RCNN-SAT architecture may be required to enable symmetry perception (e.g., the usage of LSTM recurrent connections) and also changes in the training procedure (e.g., training in a richer dataset than ImageNet that requires taking into account long-range dependencies).

Additionally, note that this work focuses solely on the detection of bilateral, reflectional symmetry. Further works may investigate other visual phenomena, as well as other forms and axes of symmetry, such as rotational and translational.

\section*{Conclusions}
The results presented here compare learned representations of symmetry in object recognition networks, and also DNNs that incorporate mechanisms to tackle long-range dependencies such as dilated convolutions, LSTMs, and Transformer architectures, and evaluate how well these networks learn a generalizable rule for symmetry detection. We concluded that DNNs based on {LSTMs  facilitate learning a generalized model of symmetry from a limited distribution of training data}. Our work motivates future investigations into architectures that incorporate LSTMs to facilitate learning general representations of symmetry from "in the wild" images where symmetry coincides with other visual phenomena.

\subsection*{Code and data availability}
The code to reproduce the experiments is available at \url{https://github.com/ssundaram21/symmetry}. Our datasets are available at \url{https://dataverse.harvard.edu/dataverse/symmetry}. 

A demo of experiment 1 can be found here: \url{https://tinyurl.com/symmetrydemo}.

\subsection*{Acknowledgments}
The authors wish to acknowledge and thank Prof. Pawan Sinha at the Department of Brain and Cognitive Sciences at MIT for his support and guidance.
This work has been supported by Fujitsu Limited (Contract No. 40009105) and the R01EY020517 grant from the National Eye Institute (NIH). 

\subsection*{Conflict of Interest Statement}
This study received funding from Fujitsu Limited. The funder through TS had the following involvement with the study: conception of the experiment, writing of this article and supervision of the study. All authors declare no other competing interests.

\section*{Methods}
\subsection*{Experiment Set 1}

\subsubsection*{Datasets}
We use specially designed synthetic test datasets that rigorously assess whether trained networks are capable of recognizing pixel-level symmetry -- in which the image flanks are perfectly mirrored -- for families of images that contain visual properties not apparent in the training dataset. We introduce several families of datasets, split between training and testing families to enable generalization testing. 

We train and test on subsets of the Band dataset families. All images are characterized by the size of a central uninformative band. The inclusion of a central band ensures that there are no local features (such as short horizontal segments created by the juxtaposition of identical pixels on the two sides of the axis of symmetry) that can be used to classify patterns as symmetric or non-symmetric.

In particular, for symmetric images with a bandsize $b$, the algorithm fills a $20 \times (10 - \frac{b}{2})$ matrix $L$ with random values drawn from the range $[1, 256]$ to represent the full grayscale spectrum. The matrix is then duplicated and flipped to create matrix $R$. We then create a matrix $B$ of size $20\times b$, with all values set to $128$, that represents the band. The three matrices are concatenated in the order $L + B + R$. For non-symmetric images, we generate separate random matrices of size $20\times (10-\frac{b}{2})$ and follow the band-creation and concatenation procedure described above. Each individual image is composed of $20\times 20$ pixel blocks, where the pixels within each block take the same values. The block size is adjusted for each network to suit the required input size.

\textit{Training image families.}
Our training set is composed of two families of images: one with a central band of size $4$, and the other without. In this way, we train on a limited subset of the full distribution (i.e. the full range of band sizes). Each image family is equally split between symmetric/non-symmetric samples. 
We created a training set with $4000$ total images ($2000$ from each family).  We use $90\%$ of the images for training and $10\%$ for validation to tune the hyper-parameters.
 
\textit{Testing image families.} 
Our test set for experiment set $1$ is composed of six additional image families. The first five feature band sizes $\{2, 6, 14, 16, 18\}$, and the sixth features a band size of $4$ with a lower average luminance, achieved by restricting the range of pixel values to $[0, 128]$. Note that each of these families are significantly differentiated from the training families. Different band sizes enable us to evaluate whether the networks are capable of capturing differing sized spatial dependencies, while the last dataset allows us to evaluate if the networks generalize to different luminances.

\subsubsection*{Networks}
We use 6 feed-forward convolutional DNN architectures, namely DenseNet~\cite{DenseNet}, Xception~\cite{Xception}, Inception ResNet V2~\cite{InceptionResNetV2}, Inception V3~\cite{InceptionV3}, ResNet101, and ResNet50~\cite{ResNet}.  Each of these networks was originally trained on the ImageNet dataset~\cite{ImageNet}. These models have been shown to match to a remarkable degree the neural activity and recognition accuracy in primates~\cite{yamins2016using}. We also use RCNN-SAT~\cite{rcnnSAT} as it has been shown to mimic human object recognition timing by incorporating recurrent connections, and thus improve previous brain models. We use the pre-trained models that are publicly available.

\subsubsection*{Training and Hyper-parameter Tuning}
 We train each network separately for the binary classification problem of symmetry detection. A particular image is classified as $1$ if it is symmetric and $0$ if it is non-symmetric. For all training procedures we tune the learning rate across $\{1\mathrm{e}{-2}, 1\mathrm{e}{-3}, 1\mathrm{e}{-4}, 1\mathrm{e}{-5}\}$. We use a batch size of $32$ and train for a minimum of $15$ epochs. We perform cross-dataset evaluation using the hyper-parameter set that achieved the highest validation accuracy. In the following, we describe the two different procedures used to train the networks: transfer-training and fine-tuning.
 
\textit{Transfer-Training.} For all object recognition DNNs we freeze the pre-trained layers and train only the classification head (2 fully-connected 128-node layers) of each network on the training dataset composed of images with band sizes $0$ and $4$. We then test whether a rule for symmetry is learned using the representations for object recognition. For the RCNN we run the network for eight time steps and average the aggregated outputs, as described in \cite{rcnnSAT}.

\textit{Fine-tuning.} We also train end-to-end the networks with the classification head for symmetry detection, such that the weights are initialized as the learned weights from ImageNet, and subsequently all layers are allowed to update based on the synthetic training set. Thus, we follow the same procedure as for transfer-training except that we allow training of all layers in the network.

\subsection*{Experiment Set 2}

\subsubsection*{Datasets}

We introduce more datasets to train the networks in more natural images and also to test the networks in more challenging images. Since we train the networks from scratch, we use an image size of $20\times 20$ pixels, which is the smallest possible for our datasets in order to facilitate training as fast as possible.

\textit{Training image families.}  Since we are training the networks from scratch, we use a larger number of training examples than in Experiment Set 1, namely, we use $1\mathrm{e}{5}$ training examples. We use the following families to train the networks:

\begin{itemize}
    \item[--] \emph{Band:} This is the same distributional makeup as the training dataset used in Experiment Set 1. All networks are trained on a dataset composed of symmetric and non-symmetric images with band size 0 and band size 4. 
    
    \item[--] \emph{Natural:} We additionally conduct experiments for networks trained on image crops from natural images.  We combine $176$ annotated symmetric images from the NYU Symmetry Database~\cite{ConvSymm2016} with $75$ symmetric images from the CVPR 2013 Symmetry Challenge~\cite{liu2013symmetry}, and $250$ non-symmetric images from the ImageNet database~\cite{ImageNet}. The NYU and CVPR datasets are well-known benchmark datasets for symmetry detection. Using standard data augmentation techniques (cropping, blurring, and rotations) we generate $24$ augmented variations of each raw image, leading to a total dataset size of $12000$ images. Dividing the dataset with a $90-10\%$ split yields a $10800$-image training set, and $1200$-image testing set. Sample Natural images with data augmentations applied are depicted in Figure\ref{fig:datasets}.
\end{itemize}

\textit{Testing image families.} 
All synthetic datasets contain $1\mathrm{e}{4}$ images, and the natural image dataset contains $1200$ images.

\begin{itemize}
\item[--] \emph{Band:} We test on all of the same testing families introduced in Experiment Set 1 (band sizes $\{$2, 6, 14, 16, 18, 4 \text{ (dark)}$\}$).

\item [--]\emph{Stripe:} In addition, we introduce the Stripe family of datasets to further elucidate how the distance between image flanks impacts perception of pixel-level symmetry, and thus how well the networks capture long-range relationships. A Stripe image contains one column of varying pixels on each flank; the rest of the image pixels are set to the constant value $128$ (and are thus uninformative). The dataset subcategories are differentiated by the positions of the two "stripe" columns, as shown in Figure \ref{fig:datasets}c. For a Stripe image with the left-side "stripe" at column $x \in \{2, 4, 6, 8, 10\}$ (where the left-most column is column $0$) we fill in a $20\times 20$ matrix with value $128$. We then replace columns $x$ and $20-x$ with the same randomly-generated $20\times 1$ vector (symmetric images) or different vectors (non-symmetric images). For each possible position of $x$ we generate both a symmetric and non-symmetric dataset.

\item[--] \emph{Natural Images:} We additionally test on the test set of natural crops ($1200$ images).

\item[--] \emph{Natural Mirrored dataset:} We create a dataset of images that are derived from the Natural dataset, but are mirrored to have identical left-right flanks. For each image in the $1200$ Natural test dataset described above, the right half was replaced by the mirrored left half of that image. Thus these images have the pixel distribution of natural images, but are precisely pixel-level symmetric (unlike fully natural images which symmetry is perceived even though there are variations between the flanks). Sample Natural Mirrored images are shown in Figure \ref{fig:datasets}.

\end{itemize}
\subsection*{Networks}

We explore the following architectures that incorporate mechanisms to tackle long-range dependencies, which are depicted in Figure \ref{fig:generalization}.

\textit{Dilated Convolutional Neural Networks.} It is an architecture that employs convolutions with “expanded”, i.e., upsampled, kernels~\cite{dilatedNetwork}. The dilation rate parameter $l$ indicates how much the kernel is expanded, introducing $l-1$ spaces between kernel elements. By increasing the dilation rate monotonically throughout the network layers, the receptive field of the network is expanded while maintaining the same number of parameters, thus facilitating the learning of long-range relationships.  We use $3\times 3$ kernels, $7$ layers, and a dilation rate of $4$, following the parameters used in~\cite{insideness} as we also found that this architecture works best for symmetry detection. 


\textit{Transformers.} The Transformer model uses a self-attention mechanism to learn long-range dependencies between inputs. Namely, the self-attention mechanisms take as input a set of image patches represented in an embedded space and compare them in a pairwise manner, such that relations across all pairs of image patches can be taken into account, independently on the distance between them. Then, the pairwise similarities between patches are projected into a set of vectors. The transformers stack many layers with self-attention and also standard fully connected layers possibly with skip connections. We use the popular  OpenAI's CLIP model, pretrained with $400$ million text/image pairs to predict text labels for image inputs \cite{Radford2021}.  

\textit{Convolutional LSTM~\cite{lstmNetwork}.} LSTM is a recurrent network that alleviates the well-known issues of training recurrent networks with a large number of unrolling steps. Since it is convolutional, it is suitable for vision problems. 
A recurrent network can be thought of as a feed-forward network applied repeatedly over many time steps, with shared weights between time steps. For each time step the output of the previous time step is fed as an input. In the “unrolled” version of the network, time steps are applied as subsequent layers.  Here, we stack three convolutional LSTM cells (of $64$ channels each) to better facilitate learning a multi-step visual routine for symmetry detection (this was found through an initial pilot experiment in which we assessed architectures with one, two or three cells).
In general, recurrent networks are capable of capturing long-range dependencies by breaking them up into sequences of local operations that are repeated over time. Previous research has demonstrated that for some visual problems involving long-range dependencies, a stacked ConvLSTM with several cells is capable of learning a simple visual routine that is generalizable to images outside the training distribution~\cite{insideness}.

\subsubsection*{Training and Hyper-parameter Tuning}

We train \textit{Dilated} and \textit{LSTM3} from scratch and transfer-train the \textit{Transformer} from the pre-trained architecture on $400$ million text/image pairs (we could not train it from scratch on our symmetry datasets, possibly because we did not have enough computational resources to train it with a large number of training examples). Separate experiments are conducted for training all networks on synthetic and natural image sets (as described above). For all experiments we use a $95\%/5\%$ split for training/validation. For both synthetic and natural training, we perform cross-dataset evaluation using the network and hyper-parameter set that achieved the highest validation accuracy.  

For the \textit{Dilated} and \textit{LSTM3}  networks, we test the following hyper-parameters.  The convolutional layers use zero-padding, the batch-size is $32$, and we explored learning rates \{$1\mathrm{e}{-1}$, $1\mathrm{e}{-2}$, $1\mathrm{e}{-3}$, $1\mathrm{e}{-4}$, $1\mathrm{e}{-5}$\}. For natural training, each network is trained on a $10800$ dataset of natural images, using the training/validation split. For the \textit{LSTM} we try different numbers of unrolling steps: \{$5$, $10$, $20$, $30$, $50$\}.

For the\textit{Transformer} architecture, we transfer-train using the standard procedure~\cite{Radford2021}. Namely, we train a logistic regressor from the representations of the layer before the output using the default parameters ($1000$ iterations, and regularizer parameter $C=0.316$, with the Transformer weights frozen).




%
%
%

\bibliography{biblio.bib}


\end{document}